\documentclass[]{article}
\pdfoutput=1
\usepackage{comment}
\usepackage{hyperref}
\usepackage{graphicx}
\usepackage{amsmath}
\usepackage{amssymb}
\usepackage{verbatim} %In the preamble 
\usepackage{bm}
\usepackage{lineno}
\usepackage{algpseudocode}
\usepackage{color}
\usepackage{cite}
\usepackage{setspace} 
\usepackage{placeins}
\usepackage{caption}
\usepackage{hyperref} 
\usepackage{mwe}
\usepackage{makecell}

\usepackage{pifont}% http://ctan.org/pkg/pifont
\newcommand{\xmark}{\ding{55}}%
\usepackage{mwe}

\usepackage{lineno}
\usepackage[ruled,vlined]{algorithm2e}
\usepackage{subcaption}
\usepackage{authblk}
\hypersetup{
  colorlinks=true,
  citecolor=blue,
  linkcolor=blue,
  filecolor=magenta,   
  urlcolor=cyan,
}
\title{Local Padding in Patch-Based GANs for Seamless Infinite-Sized Texture Synthesis}
\author[1]{Alhasan Abdellatif \footnote{a.abdellatif@hw.ac.uk}}
\author[1]{ Ahmed H. Elsheikh\footnote{a.elsheikh@hw.ac.uk}}
\author[1]{ Hannah P. Menke\footnote{h.menke@hw.ac.uk}}

\affil[1]{Heriot-Watt University}

\date{}

\begin{document}
%\linenumbers
\maketitle
\sloppy

\begin{abstract}
Texture models based on Generative Adversarial Networks (GANs) use zero-padding to implicitly encode positional information of the image features. However, when extending the spatial input to generate images at large sizes, zero-padding can often lead to degradation in image quality due to the incorrect positional information at the center of the image. Moreover, zero-padding can limit the diversity within the generated large images. In this paper, we propose a novel approach for generating stochastic texture images at large arbitrary sizes using GANs based on patch-by-patch generation. Instead of zero-padding, the model uses \textit{local padding} in the generator that shares border features between the generated patches; providing positional context and ensuring consistency at the boundaries. The proposed models are trainable on a single texture image and have a constant GPU scalability with respect to the output image size, and hence can generate images of infinite sizes. We show in the experiments that our method has a significant advancement beyond existing GANs-based texture models in terms of the quality and diversity of the generated textures. Furthermore, the implementation of local padding in the state-of-the-art super-resolution models effectively eliminates tiling artifacts enabling large-scale super-resolution. Our code is available at \url{https://github.com/ai4netzero/Infinite_Texture_GANs}.

\end{abstract}

\section{Introduction}
Texture synthesis refers to the process of generating arbitrary-size textures that are visually similar to a given input example, while also being diverse and not a simple duplication of the input. This problem has numerous applications in fields such as gaming, virtual reality and graphic design, where high-quality textures are critical for creating realistic and visually appealing environments. The use of Generative Adversarial Networks (GANs) \cite{goodfellow2014generative} in generating realistic textures has been widely explored. Several methods have been proposed for this task, including Spatial GAN \cite{jetchev2016texture}, PSGAN \cite{bergmann2017learning}, adversarial expansion \cite{zhou2018non} and SinGAN \cite{shaham2019singan}. However, using these models to generate arbitrary large-scale textures while maintaining their quality and diversity is not trivial.

Increasing the input latent space size in these models is limited by GPU memory, which restricts the size of the output image. In addition, when using zero-padding, extending the input dimensions results in degraded quality due to the propagation of incorrect positional information. Furthermore, zero-padding can cause limited variability at the corners of the generated images and can often result in tiling or seaming artifacts between the generated patches in incremental generation, i.e., patch-by-patch generation. This problem is more pronounced in image translational tasks, where several works have tried to alleviate the discontinuities at the patch borders either by post-training tiling \cite{de2018stain} or integrating the tiles into training \cite{bohland2023improving}.

To address these challenges, we propose a patch-based GAN model with a novel padding method that is capable of synthesizing stochastic textures of infinite size and that is trainable on a single texture image. Because of the self-similarity and homogeneity within texture images, it is sufficient to maintain their structure by sharing local information only. Relying on this concept, our model uses \textit{local padding}, instead of zero-padding, where the inputs to the generator's convolutional layers are padded with content from neighbouring patches to ensure seamless concatenation. During training, the model generates small, locally correlated patches, which it learns to seamlessly stitch together using padded information. Using incremental generation, at the inference time, we can extend the generation to arbitrary large sizes while maintaining the texture structure and diversity. 

The main contributions of the paper:
\begin{itemize}
	\item We propose \textit{local padding} as a new way of padding the inputs before convolutional operations that allows seamless patch-by-patch texture synthesis.
	
	\item We demonstrate that our trained models are capable of generating higher-quality texture images than existing models while maintaining the fine details and variability exhibited by the original examples and that they are scalable to any resolution by incremental generation. % compared to others?
	
	\item We also show that local padding can be used in the state-of-the-art super-resolution models, such as Real-ESRGAN \cite{wang2021real}, to avoid tiling artifacts when super-resolving large inputs.
\end{itemize}

The paper is organized as follows: in section \ref{back}, we present a review of related work. In section \ref{method}, we discuss the proposed patch-by-patch generation with local padding. In section \ref{results}, we present several applications of the developed method and discuss the results. Finally, the conclusions of this manuscript are presented in section \ref{concl}.

\section{Literature Review}

\label{back}
%\subsection*{Generative Adversarial Networks (GANs)}
\textbf{GANs for texture synthesis.} Generative Adversarial Networks (GANs) \cite{goodfellow2014generative} have gained significant attention in recent years due to their ability to generate realistic images. They have found numerous applications in computer vision, ranging from image-to-image translational tasks \cite{zhu2017unpaired,tian2018cr,zhao2020towards}, super-resolution \cite{ledig2017photo,wang2021real,zhang2021designing}, and image in-painting \cite{pathak2016context,yu2018generative}. GANs have also shown great potential for generating realistic textures, where the challenge is to generate samples of arbitrarily large sizes while preserving the coherence and consistency of the given example. Spatial GAN \cite{jetchev2016texture} builds upon the DCGANs architecture \cite{radford2015unsupervised} by transforming the generator and discriminator into fully convolutional networks, where the output texture image can be expanded in size by expanding the spatial input. Periodic Spatial GAN (PSGAN) \cite{bergmann2017learning} proposes to generate textures with periodic patterns by incorporating a periodic input into the generator network. Adversarial expansion \cite{zhou2018non} trains a GAN model to double the size of the input texture image. However, expanding the input size of the latent space leads to incorrect positional encoding in the generated images and hence degrades the quality. Moreover, adversarial expansion models do not parametrize the stochasticity of the texture, and instead performed diversification by shuffling and cropping.

SinGAN \cite{shaham2019singan} trains multi-scale generators to generate realistic images, including textures, from a single input image. TileGAN \cite{fruhstuck2019tilegan} designed a tiling framework to synthesize large-scale texture images based on a neighbourhood similarity search. While these models successfully generated high-resolution images of textures, the zero-padding used leads to limited spatial variability around the boundaries of the generated images in case of SinGAN and visible artifacts between the tiles with TileGAN. In addition, the TileGAN method requires storing latents representations of a large number of generated examples to be searched for similarity matching, which is time-consuming. 

\textbf{Incremental Generation.} In patch-by-patch generation, the model synthesizes one small patch at a time, then correlated patches are assembled to form a larger image. This allows the model to generate images with an infinite size and avoid the problem of limited resources and the training instabilities associated with generating a large image in a single forward pass \cite{brock2018large}. COCO-GAN \cite{lin2019coco} trains a GAN model that conditions image patches on coordinates and then assembles patches that share a global latent vector. This allows for limited extrapolation of the images by extending the coordinates. InfinityGAN \cite{lin2021infinitygan} then extended the method to natural images by employing a padding-free generator. The authors removed zero-padding in their generator and instead padded the inputs with neighbouring content, which allowed for seamless concatenation. To model the position of the patches, (e.g., sky, land), they employed an implicit neural function with CoordConv \cite{liu2018intriguing}, where the hidden representations are concatenated with positional embeddings. 

LocoGAN \cite{struski2022locogan} trains fully convolutional GANs to generate sub-images instead of the full image and uses coordinates to inform the model which part of the image is being generated. The main drawback of their model is that it is limited to periodic texture due to the nature of the periodic coordinates used. ALIS \cite{skorokhodov2021aligning} used a spatially-equivariant generator where they modified the AdaIN algorithm \cite{huang2017arbitrary} such that the modulating parameters are spatially-interpolated. This approach enabled the generation and assembly of vertical patches of natural scenes. However, the model suffers from content repetition when the global anchors do not change fast enough to allow for variations. In addition, the padding used in the generation leads to blocky artifacts and discontinuity between the generated patches. Unlike natural images that require global coordination between the different patches, texture images exhibit a high degree of locality and self-similarity. Applying local padding at every layer enables our model to capture local texture structure without explicit global coordination between patches.

\section{Methodology}
\label{method}
\begin{figure*}[!t]
	\centering
	\includegraphics[scale=0.364]{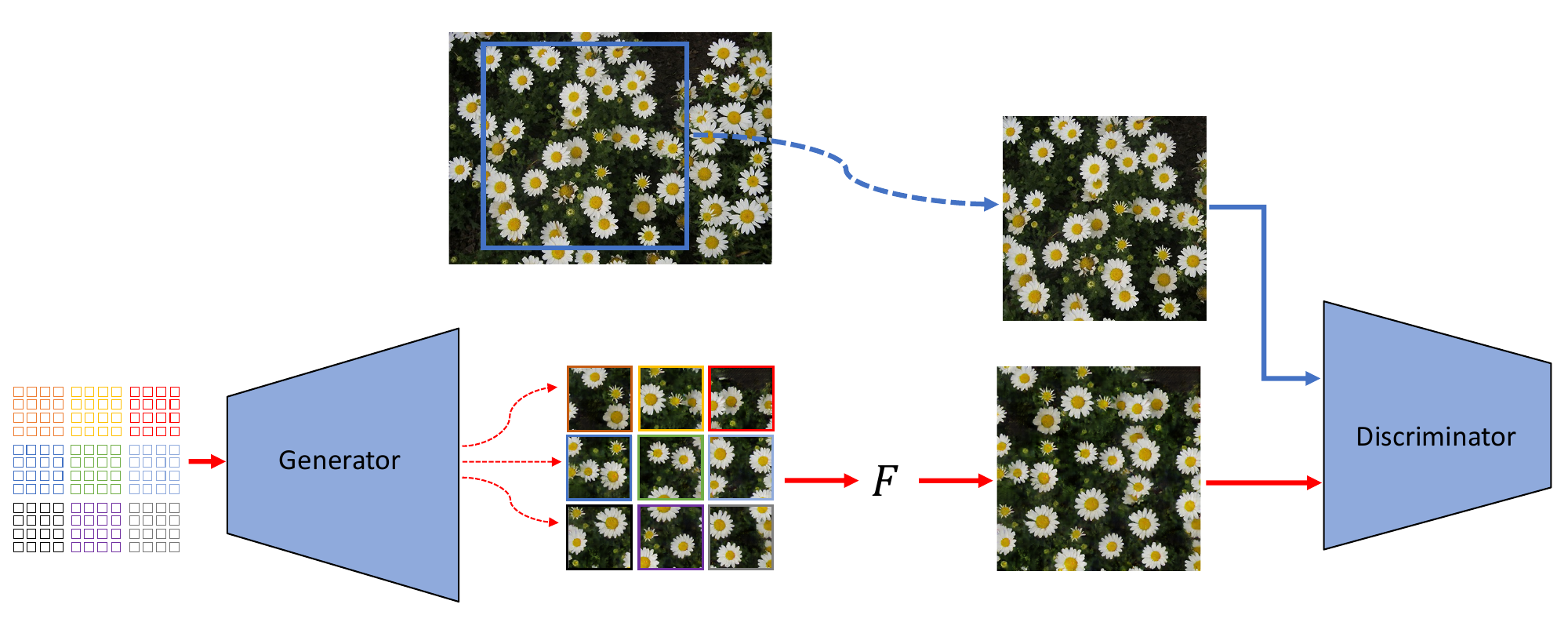}
	\caption{Overview of the training process for the proposed method. An image patch is cropped randomly from the single texture image and is passed to the discriminator. The generator takes $3\times 3$ random spatial inputs $z_i$, each of size $4 \times 4$, to generate, in parallel, $3\times3$ small-size patches $x_i = G(z_i)  \, \, \forall \, i \in 1 \dots 9$ that are passed to an assembling function $F$ to form a larger image $X = F(x_1,\dots,x_9)$. The assembled image is then passed to the discriminator for evaluation. The blue arrows indicate the real patches path while the red arrows indicate the generated patches path.}
	\label{fig:method_overview}
\end{figure*}

The proposed method relies on homogeneity and self-similarity within texture images. This property allows us to synthesize texture patches using only information from the neighbouring patches. An overview of the method is presented in Figure \ref{fig:method_overview}. Following \cite{jetchev2016texture}, both the generator and discriminator are fully convolutional neural networks. Multiple patches are randomly cropped from the single texture image and are fed into the discriminator. Unlike traditional approaches that expand spatial noise to generate larger images, we limit the generator network to simultaneously produce, $N\times N$ small-size patches, each patch is of size $h \times h$. In the demonstration Figure \ref{fig:method_overview}, $N$ and $h$ are set to 3 and 128, respectively. The model learns to seamlessly assemble these patches into one image $X = F(x_1,\dots,x_{N^2})$, where $F$ is a simple concatenation function. The assembled image is then passed to the discriminator. The seamless assembly is facilitated by shared information between patches, achieved through a novel padding technique we call \textit{local padding}.

\textbf{Local Padding.} Since most texture images are stationary and homogenous, capturing the local spatial structure can be achieved by sharing border features of the patches. Therefore, we pad the input to the convolutional layers in the generator with content from the neighbouring patches, instead of the conventional zero-padding. We introduce \textit{local padding} as a type of padding used in patch-based generation, where the inputs to the convolutional operations at all levels in the generator are padded with the boundary content of neighbouring patches.

\begin{figure}[!t]
	\begin{subfigure}{0.5\textwidth}
		\centering
		\includegraphics[width=0.93\linewidth]{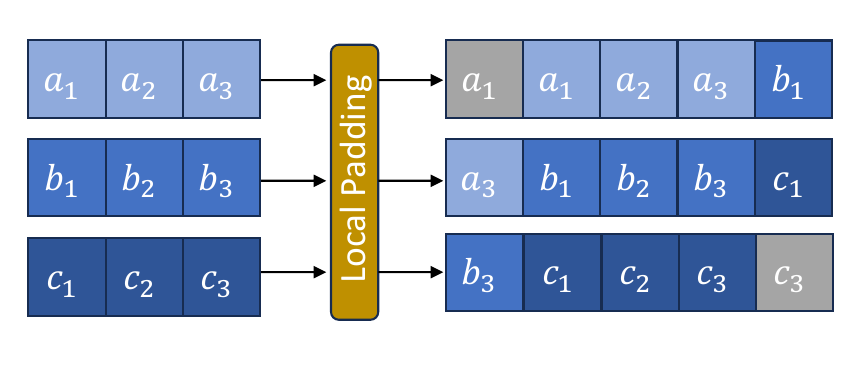}
		\caption{Local Padding}
	\end{subfigure}
	\begin{subfigure}{0.5\textwidth}
		\centering
		\includegraphics[width=0.93\linewidth]{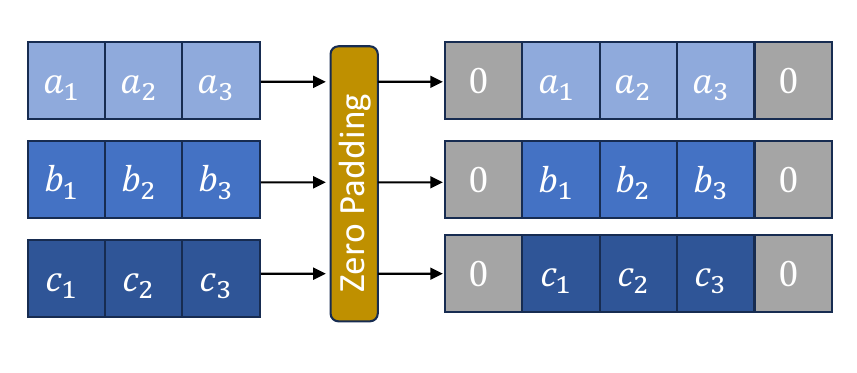}
		\caption{Zero Padding}
	\end{subfigure}
	\caption{Comparison between local padding and zero-padding applied to a batch of 3 inputs, each of size $1 \times 3$.  The output image patches from the 3 inputs are assembled together horizontally. a) In local padding, the features of neighbouring patches are used to pad each input before passing to a convolutional layer. Replicate padding is used for the border pixels in the outer patches as there is no neighbouring content. b) Zero-padding simply pads each input with zero values, leading to border inconsistency when assembling the patches together.}
	\label{fig:explain_local_pad}
\end{figure}

Typically, padding involves adding extra rows and columns of zeros around the input to a convolutional layer to ensure that the convolutional filters can be applied to the edges. However, using zero-padding in a patch-by-patch generator results in visible seams between the concatenated patches. This mismatch occurs because the pixels at the boundaries may not align perfectly with those in the neighbouring patches. Local padding addresses this issue by filling in the padded pixels with values taken from the layer input of the neighbouring patches, as depicted in Figure \ref{fig:explain_local_pad}. The figure shows a 1-dimensional generation problem, where the rows on the left side represent inputs corresponds to 3 different image patches $\{x_A,x_B,x_C\}$. that are assembled together horizontally to form one image $F(x_A,x_B,x_C)$. Local padding is applied to the inputs of all convolutional layers in the generator network.

Global operations such as batch-normalization and nearest neighbour up-sampling are still performed on each patch independently. This ensures that the convolutional filters can be applied to the edges of the feature maps without losing border information. In addition, sharing the padding between the patches and assembling them together during training allows the model to learn to seamlessly stitch the patches based on the shared padding. Padding from neighbouring patches is similar to \cite{lin2021infinitygan}, however we perform the padding at all layers in the generator not just at the input layer. This provides the neighbouring patches context for all levels when generating the local patch and passes the positional information without the need for explicit coordination.

The amount of padding applied depends on the size and the number of the convolutional filters used. For example, a $4\times4$ input is padded with 1 value at both dimensions to be of size $6\times6$ before passing it to a $3\times3$ convolution. For the outer patches, we use replicate padding to extend the input along the edges, as we do not have neighbouring patches to provide padding values. This approach lends itself to an easy extension to infinite image generation as we will detail next.

\begin{figure}[!t]
	\centering
	\includegraphics[width=\linewidth]{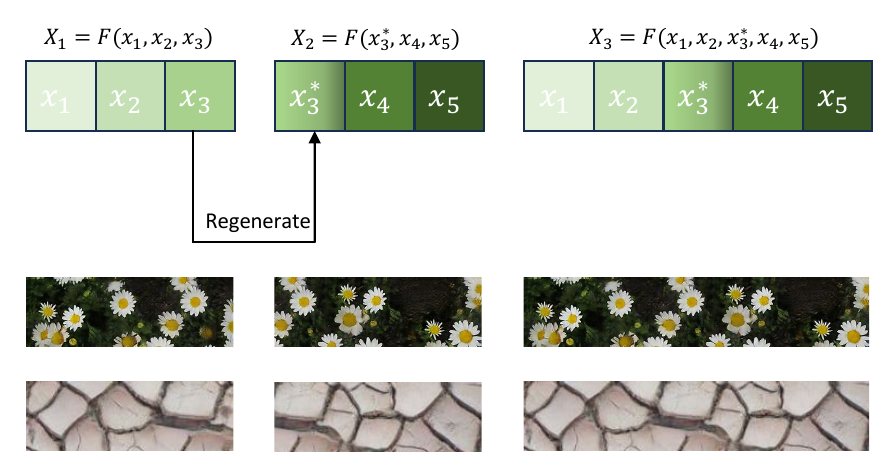}
	
	\caption{Scaling in the horizontal direction: (Top) Two images $X_1$ and $X_2$, each composed of $1\times 3$ generated patches, where the rightmost patch in $X_1$ is regenerated to match the interior patches of $X_2$ (i.e., $x_3$ is regenerated to be consistent with $x_4$). $X_3$ is then formed by concatenating $X_2$ with $X_1$ patches after dropping $x_3$. (Bottom) Two examples of generated texture demonstrate the horizontal scaling process, where the rightmost part of the first column has been regenerated with local padding to match the parts with the images in the second column.}
	\label{fig:explain_scale}
\end{figure}

\textbf{Scaling to Infinite Sizes.} We explain the scaling process for 1-dimensional patches in the horizontal direction in Figure \ref{fig:explain_scale}. First, using patch generation with local padding, we can generate an image $X_1$ which is a concatenation of the patches $x_1,x_2$ and $x_3$. The same process is repeated to generate a new image $X_{2}$. However, the rightmost patch in $X_1$ (i.e., $x_3$), which was generated using replicate padding, will be regenerated in $X_{2}$ using local padding, i.e., by padding from features in $x_2$ and $x_4$, so that the regenerated patch $x_3^*$ is consistent with both images. The patch $x_3$ is then dropped and the remaining patches are concatenated with the patches in $X_{1}$ to form a larger image $X_{3} = F(x_1,x_2,x_3^*,x_4,x_5)$. In the last two rows of the figure, we show examples in which the leftmost patch in $X_2$ is similar to the rightmost patch in $X_1$, except that the right side has been modified with local padding to match the interior patches in $X_2$. This incremental process can be repeated in the two dimensions to generate images of infinite sizes while maintaining constant GPU memory and avoiding visible seams or artifacts between the patches. 

\section{Experiments}
\label{results}
\subsection{Experiment Setup}

%\textbf{Datasets.}
The training examples used in the experiments were obtained from the supplementary materials in \cite{zhou2018non}. We selected the images such that they are homogeneous and stationary in nature. This is because the developed patch-by-patch approach cannot handle non-stationary patterns.
% why?

\begin{figure}[!t]
	%\centering
	\begin{subfigure}{1\textwidth}
		\centering
		\includegraphics[scale = 0.75]{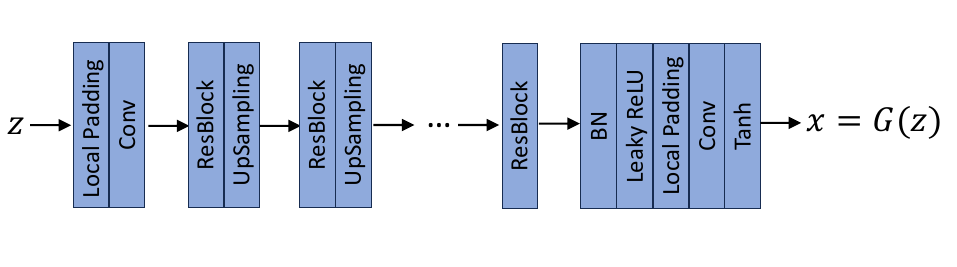}
		\caption{Generator Architecture.}
	\end{subfigure}
	\begin{subfigure}{1\textwidth}
		\centering
		\includegraphics[scale = 0.75]{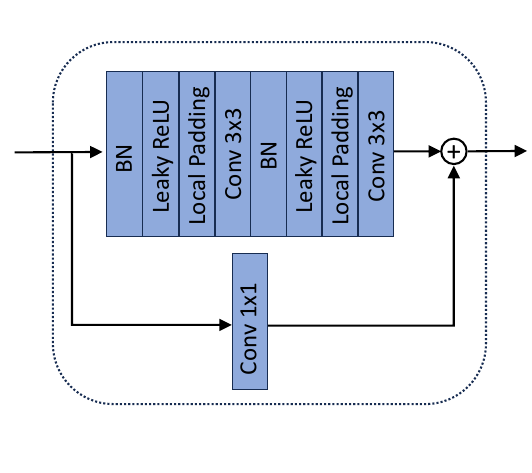}
		\caption{Residual Block}
	\end{subfigure}
	\caption{The architecture of the generator network. The generator takes a spatial noise input $z$ and successively upsamples it to generate an image patch $x = G(z)$. Each $3\times3$ convolutional layer is preceded by local padding where the features are padded with content from neighbouring patches. Batch normalization and upsampling operations are performed on each patch independently.}
	\label{fig:generator_arch}
\end{figure}

%\noindent \textbf{Implementation Details.}
The generator network is built using ResNets blocks \cite{he2016deep} with batch-normalization and nearest neighbour upsampling operations. In addition, we modified the generator to be a fully convolutional network similar to \cite{jetchev2016texture,bergmann2017learning}, removed all zero-padding following \cite{lin2021infinitygan}, and used local padding before every convolutional layer instead. The architecture of the generator is shown in Figure \ref{fig:generator_arch}, where the spatial input $z$ is passed through successive residual blocks interleaved by upsampling layers. We used a PatchGAN discriminator \cite{isola2017image,park2019semantic}, which is designed to provide a more fine-grained evaluation of the generated images by focusing on the local features of the image rather than the global features. Similar to \cite{bergmann2017learning}, we removed the normalization layers in the discriminator and found this to be more stable and boosted performance when training on a single image. The GANs model is trained with the non-saturating logistic loss with spectral normalization \cite{miyato2018spectral} applied to the discriminator weights. The learning rate of both the generator and discriminator is set to 0.0002 and Adam \cite{kingma2014adam} is used for optimization with $\beta_1 = 0$ and $\beta_2= 0.999$.

During training, we set $N = 3$. While other combinations are possible, $3\times 3$ is the simplest combination in which the generator can learn the local information between patches (i.e., 1 central and 8 neighbouring patches). During inference, one can set $N$ to be larger to maximize GPU utilization. The choice of the cropping size depends on the size of the training image, with larger cropping size leading to less diversity, since we are training on a single image. Moreover, the receptive field (RF) of the PatchGAN discriminator is selected to be larger than the size of the texture objects presented in the image so that the model can evaluate them. We present in Table \ref{tab:hyperparm}, the hyper-parameters used in some of the experiments, where the first column refers to the training image name in the supplementary materials of \cite{zhou2018non}.

\begin{table*}
	%\begin{center}
	%\scriptsize
	\centering
	\begin{tabular}{|c|c|c|c|c|}
		\hline
		Image Name (size) & Random Crop	 & Number of layers in G & Number of layers in D \\
		\hline
		12 ($450\times 600$)    & 128       &5          & 3          \\
		\hline
		34   ($450\times 600$) &128       & 5          & 4          \\
		\hline
		241 ($440\times614$) &192       & 6          & 4          \\
		\hline
		73   ($400\times 600$)  & 128       &5          & 3          \\
		\hline
		417 ($192\times 192$)    & 48       &4          & 3          \\
		\hline
	\end{tabular}
	\caption{Hyper-parameters selected for some experiments. Each row represents a distinct image experiment with the corresponding chosen parameters.}
	\label{tab:hyperparm}
%\end{center}
\end{table*}

\subsection{Results}
\begin{figure*}
	\centering
 	\includegraphics[scale=1]{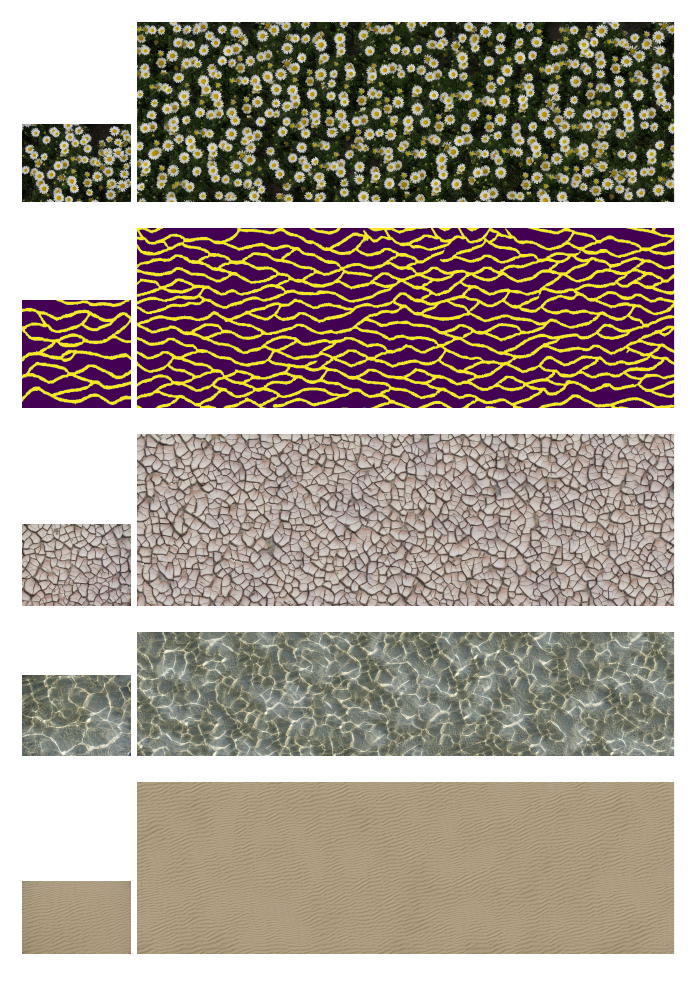}
	\caption{Examples of texture images generated by the proposed method (right column) given the source texture (left column). From top to bottom, the image names are 241, 10, 12, 34 and 73.}
	\label{fig:res_examples}
\end{figure*}

\begin{figure*}
	\centering
 	\includegraphics[scale=1]{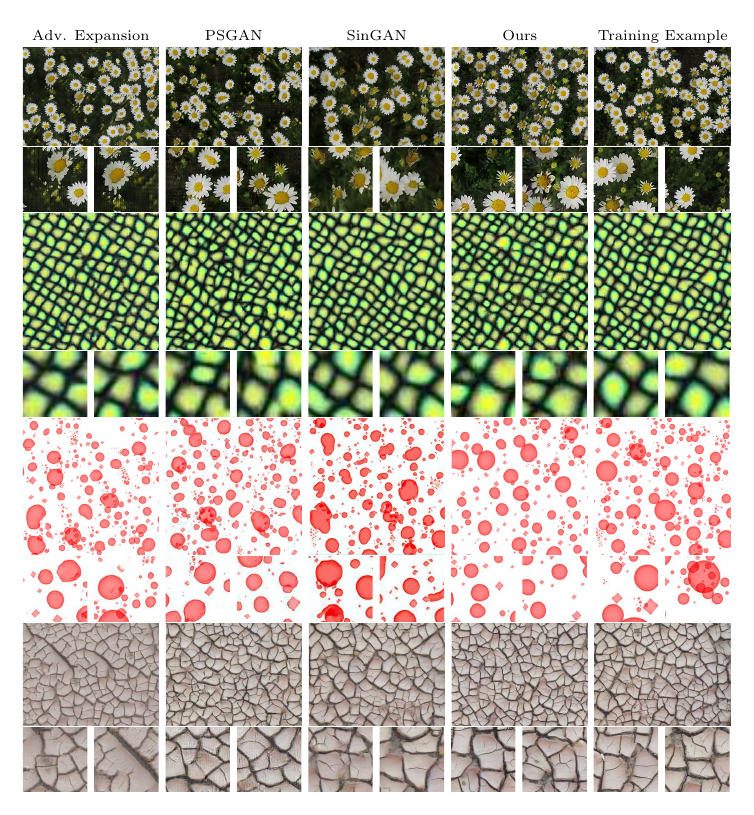}
	\caption{Qualitative comparison between texture generation models based on GANs. For each training example, we show the generated images followed by the corresponding patches for closer examination. From top to bottom, the image names are 241, 417, 221 and 12.}
	\label{fig:visual_compare}
\end{figure*}

First, we present the visual quality of some generated textures by the proposed method in Figure \ref{fig:res_examples}. As observed, the method generates textures with fine details, preserving the overall structure and diversity of the original examples. In Figure \ref{fig:visual_compare}, we compare the proposed approach against a number of published methods including: Adversarial Expansion \cite{zhou2018non}, PSGAN \cite{bergmann2017learning}, and SinGAN \cite{shaham2019singan} in terms of the quality of the generated texture. As shown, both adversarial expansion and PSGAN generate texture of lower quality since expanding the spatial input of the model changes the positional encoding.

\begin{figure*}
	\centering
	\includegraphics[scale=0.9]{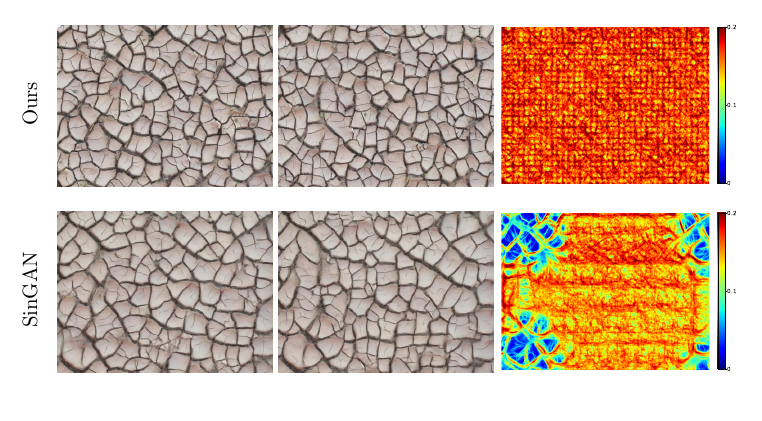}
	\caption{Diversity comparison between our model and SinGAN. The first two columns show two examples generated by our trained model and SinGAN, the last column shows the per-pixel standard deviation computed over 50 samples.}
	\label{fig:std_compare}
\end{figure*}

While SinGAN generates reasonable results, it tends to produce smooth images due to the multi-scale training scheme. As a result, some of the high-frequency details in the original example are lost, as shown in the last row. Moreover, because of the zero-padding used in SinGAN, the generated examples exhibit limited variability around the boundaries as shown in Figure \ref{fig:std_compare}, where we plot the standard deviation of 50 generated samples computed per pixel.

\begin{figure*}
	\centering
	\includegraphics[width=0.7\linewidth]{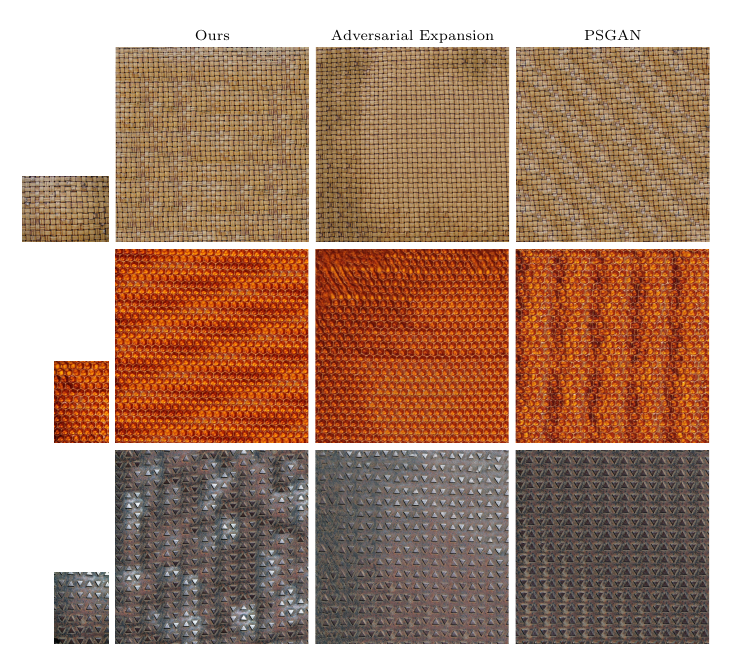}
	\caption{Regular textures generated by the proposed method with periodic inputs, adversarial expansion and PSGAN. While the stochastic variations of Adversarial Expansion and PSGAN are low across the spatial domain, our models were able to generated images that maintained the regular patterns in the original examples while also being diverse. From top to bottom, the image names are 182, 214, and 215.}
	\label{fig:regular_texture}
\end{figure*}

To generate regular texture, we incorporated periodic inputs proposed in PSGAN in the latent space of our models. In Figure \ref{fig:regular_texture}, we compare between our method with periodic input, Adversarial Expansion and PSGAN based on generated regular textures. As shown adversarial expansion tries to duplicate the structure presented in the original examples with minimal stochastic variations across the spatial domain. In addition, the zero-padding creates boundary artifacts in the generated textures. PSGAN tends to generate repetitive patterns, diminishing its capacity to produce diverse outputs. On the other hand, the proposed method was able to maintain the regularity of the texture as well as generate diverse and stochastic outputs.

Table \ref{sifid} presents a quantitive comparison using SIFID (Single Image FID) \cite{shaham2019singan}, a metric used to assess the quality and the diversity of the generated images by computing the FID distance between feature statistics of the real image and those of the generated samples. Average SIFID values are calculated over 50 images for PSGAN \cite{bergmann2017learning}, SinGAN \cite{shaham2019singan}, and our method. We also show, in Table \ref{fid}, FID values calculated for random image patches selected from the real image and the generated images. The results show that the texture images generated by our method consistently achieve lower SIFID and FID values across all cases, indicating superior quality and diversity compared to those produced by PSGAN and SinGAN. SinGAN models, in particular, tend to have large SIFID values due to limited variability around the corners of the generated images. Finally, in Table~\ref{tab:texture-gans}, we compare between the developed method and other related GANs work based on GPU scalability, the use of coordinates, the use of zero-padding, and whether they are trainable on single images.

\begin{table}
	
	\begin{center}
		
		\begin{tabular}{|c|c|c|c|}
			\hline
			Image & PSGAN & SinGAN & Ours \\
			\hline
			12   &0.22       & 0.27         & \textbf{0.04}        \\
			\hline
			34   &0.08       & 0.76         & \textbf{0.07}     \\
			\hline
			241   &0.12       & 0.69     & \textbf{0.09}        \\
			\hline
			221   & 0.64      & 0.43     & \textbf{0.30}        \\
			\hline
			417   &0.51      & 0.18     & \textbf{0.12}        \\
			\hline
			
		\end{tabular}
	\end{center}
	\caption{Comparison of Single Image FID (SIFID) Values for Different methods (lower is better).}
	\label{sifid}
\end{table}

\begin{table}
	
	\begin{center}
		
		\begin{tabular}{|c|c|c|c|}
			\hline
			Image & PSGAN & SinGAN & Ours \\ \hline
					12 & 0.04 & 0.03 & \textbf{0.01}\\ \hline
					34 & 0.02 & 0.05 & \textbf{0.01} \\ \hline
					241 & 0.05& 0.11 & \textbf{0.04} \\ \hline
					221 & 0.03 & 0.05 & \textbf{0.02} \\ \hline
					417 & 0.07 & \textbf{0.02} & \textbf{0.02} \\ \hline		
		\end{tabular}
	\end{center}
	\caption{Comparison of FID Values computed for Images Patches (lower is better).}
	\label{fid}
\end{table}

\begin{table*}
	%\begin{center}
	\scriptsize
	\centering
	\begin{tabular}{|c|c|c|c|c|}
		\hline
		GANs Method & \makecell{Constant GPU \\ Scalability} & Coordinate-free & \makecell{Avoids \\ zero-padding} & Single-image \\
		\hline
		PSGAN \cite{bergmann2017learning} & \xmark & \checkmark & \xmark & \checkmark \\
		\hline
		SinGANs \cite{shaham2019singan} & \xmark & \checkmark & \xmark& \checkmark\\
		\hline
		Adversarial Expansions \cite{zhou2018non} & \xmark & \checkmark & \xmark & \checkmark \\
		\hline
		LocoGAN \cite{struski2022locogan} & \xmark & \xmark & \xmark & \checkmark \\
		\hline
		ALIS \cite{skorokhodov2021aligning} & \checkmark & \xmark & \xmark & \xmark \\
		\hline
		InfinityGAN \cite{lin2021infinitygan} & \checkmark & \xmark & \checkmark & \xmark \\
		\hline
		Ours & \checkmark & \checkmark & \checkmark& \checkmark\\
		\hline
	\end{tabular}
	%\end{center}
	\caption{Comparison of various GANs methods focusing on GPU scalability, coordinate-free implementation, padding-free processing, and suitability for generating single images.}
	\label{tab:texture-gans}
\end{table*}

\textbf{Local Padding in Super-Resolution models.} 
Super-resolution based on deep learning models has been an active area of research in the last few years, where the models learn to reproduce the high-frequency details lost in compressed, noisy or blurry images. However, super-resolving large images is limited by the GPU memory, and hence tiling is often used where the large input image is broken down into smaller overlapping tiles or patches. Each tile is processed independently, and the outputs are stitched back together to form the final large image. However, seaming artifacts can appear when assembling the patches together due to mismatches around the boundaries. To address this problem, we apply local padding in the state-of-the-art Real-ESRGAN \cite{wang2021real} model.

\begin{figure}
	\centering
	\includegraphics[width=\linewidth]{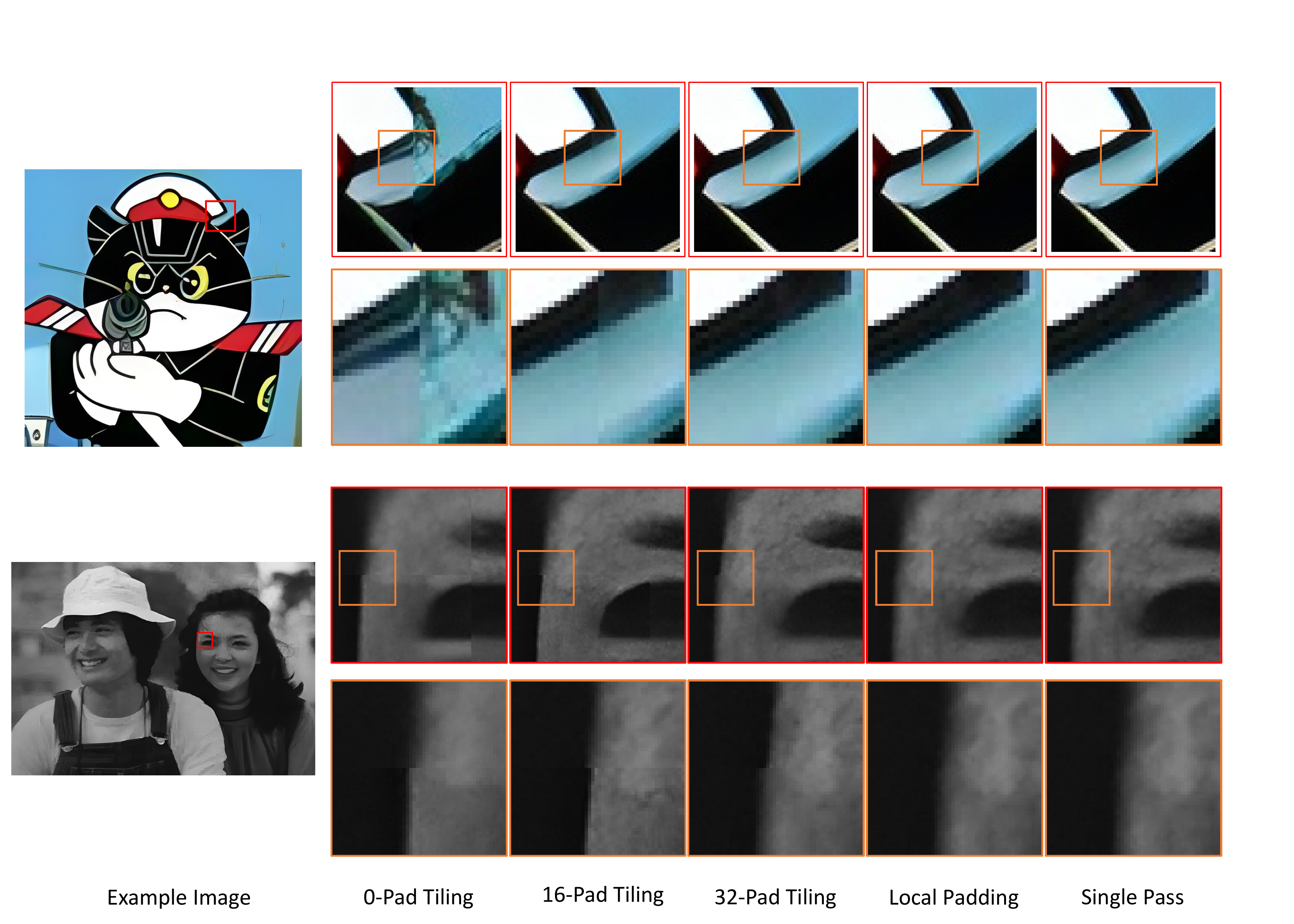}
	\caption{Applications of local padding in super-resolution. Two examples of super-resolved images using Real-ESRGAN model. The input images are processed using tiling with different overlapping sizes and local padding. The results are compared to passing the inputs to the model in a single shot.}
	\label{fig:sr_example}
\end{figure}

Since no global operations is used in the Real-ESRGAN generator (e.g., no batch normalization), we can apply the method directly to the pre-trained model by dropping all the zero paddings and instead use local padding. In Figure \ref{fig:sr_example}, we used Real-ESRGAN to super-resolve images using both tiling with different overlapping sizes and local padding. Although increasing the overlapping size between the patches smooths down the discontinuities, the cutting lines are inevitable. On the other hand, local padding resulted in no discontinuity between the patches and produced details similar to the single forward pass to the model.

\begin{figure}
	\centering
	\includegraphics[width=\columnwidth]{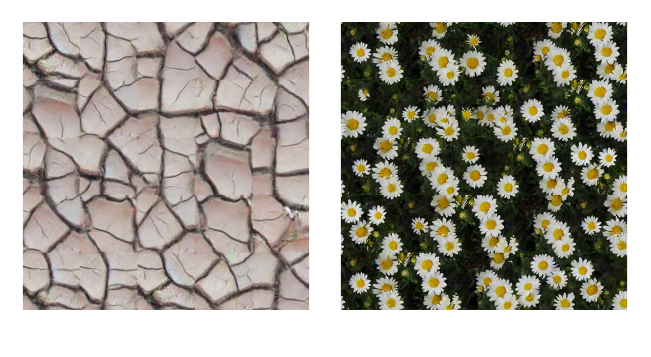}
	\caption{Two examples of texture images generated using zero-padding instead of local padding. The images illustrate how zero-padding leads to visible seaming artifacts between the patches.}
	\label{fig:zeropad example}
\end{figure}

\textbf{Ablation study.} To evaluate the effectiveness of the proposed method, we conducted an ablation study where we ablated the local padding and replaced it with the conventional zero-padding. Figure \ref{fig:zeropad example} shows examples of images generated using zero-padding where noticeable seams and discontinuities become apparent between patches, disrupting the visual coherency of the generated output. In contrast, the utilization of local padding in the generator effectively mitigates these issues, resulting in a visually consistent and coherent image as shown in Figure \ref{fig:res_examples}.

\begin{figure}[h]
	\centering
	\includegraphics[width=\linewidth]{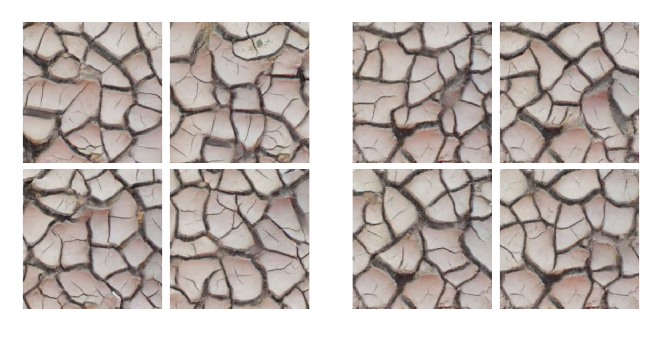}
	\caption{Comparison of texture generated using a fully convolutional generator (left) versus a generator with a fully connected layer (right). The fully convolutional generator demonstrates diverse texture samples, whereas the generator with a fully connected layer suffers from spatial mode collapse, producing less varied textures, as evidenced in areas such as the bottom left corners.}
	\label{fig:noFCG}
\end{figure}

We also studied the effect of having a fully convolutional generator instead of a traditional generator that uses a fully connected input layer followed by convolutional layers similar to \cite{brock2018large}. The results in Figure \ref{fig:noFCG} show that the fully convolutional generator is able to generate diverse and visually appealing textures, while the generator with a fully connected layer suffers from spatial mode collapse, producing less varied textures.

\section{Conclusion}
\label{concl}
We have presented a novel approach for synthesizing textures of infinite size, trained on a single image. The proposed patch-based generation with local padding addresses the limitation of memory scalability and the generation of high-quality, diverse, large size textures, that have challenged previous methods. The trained models can successfully generate visually appealing texture images with intricate details and seamless transitions between patches. Large-scale textures can also be synthesized incrementally in a scalable mannar without a proportional growth in GPU memory usage. Nevertheless, this approach has its limitations, including the requirements to cache a fraction of the feature maps in the scaling steps and the inability to handle non-stationary textures. Future work could address the latter problem by incorporating global inputs to capture long term relationships in non-stationary patterns. 

\section{Acknowledgment}
This work was partially funded by the EPSRC grant reference EP/Y006143/1. The first author acknowledges TotalEnergies for supporting the early part of this work conducted during his PhD studies at Heriot-Watt University.

\bibliographystyle{ieeetr}
\bibliography{paper3}

\end{document}